%% file: aaai24.tex
\documentclass[letterpaper]{article} % DO NOT CHANGE THIS
\usepackage[table,xcdraw]{xcolor}
\usepackage[submission]{aaai24}  % DO NOT CHANGE THIS
\usepackage{times}  % DO NOT CHANGE THIS
\usepackage{helvet}  % DO NOT CHANGE THIS
\usepackage{courier}  % DO NOT CHANGE THIS
\usepackage[hyphens]{url}  % DO NOT CHANGE THIS
\usepackage{graphicx} % DO NOT CHANGE THIS
\usepackage{natbib}  % DO NOT CHANGE THIS AND DO NOT ADD ANY OPTIONS TO IT
\usepackage{caption} % DO NOT CHANGE THIS AND DO NOT ADD ANY OPTIONS TO IT
\usepackage{algorithm}
\usepackage{algorithmic}

\usepackage[hyphens]{url}
\usepackage{minitoc}
\usepackage{natbib}
\usepackage{subcaption}
\usepackage{dsfont} % indicator function
\usepackage{amssymb}
\usepackage{amsthm}
\usepackage{stmaryrd} % llbracket
\usepackage{times}
\usepackage{epsfig}
\usepackage{graphicx}
\usepackage{amsmath}
\usepackage{dsfont}
\usepackage{amssymb}
\usepackage{booktabs}
\usepackage{multirow}
\usepackage{pifont}% http://ctan.org/pkg/pifont
\usepackage{comment}
\usepackage{caption}
\usepackage{tikz}
\usepackage{comment}
\usepackage{amsmath,amssymb} % define this before the line numbering.
\usepackage{color}
\definecolor{Gray}{gray}{0.85}
\usepackage{array}
\usepackage{eqparbox}

\usepackage{graphics}
\usepackage{graphicx}
\usepackage[utf8]{inputenc} % allow utf-8 input
\usepackage[T1]{fontenc}    % use 8-bit T1 fonts
\usepackage{url}            % simple URL typesetting
\usepackage{booktabs}       % professional-quality tables
\usepackage{amsfonts}       % blackboard math symbols
\usepackage{nicefrac}       % compact symbols for 1/2, etc.
\usepackage{microtype}      % microtypography
\usepackage{multirow}
\usepackage{amsfonts}
\usepackage{amsmath}
\usepackage{amssymb}
\usepackage{newfloat}
\usepackage{listings}
\DeclareCaptionStyle{ruled}{labelfont=normalfont,labelsep=colon,strut=off} % DO NOT CHANGE THIS
\floatstyle{ruled}
\newfloat{listing}{tb}{lst}{}
\floatname{listing}{Listing}
\title{SAfER: Layer-Level Sensitivity Assessment for Efficient and Robust Neural Network Inference}
\author{
    Edouard Yvinec$^{1,2}$ , Arnaud Dapogny$^2$ , Kevin Bailly$^{1,2}$
}

\affiliations{
    Sorbonne Université$^1$, CNRS, ISIR, f-75005, 4 Place Jussieu 75005 Paris, France \\
  Datakalab$^2$, 114 boulevard Malesherbes, 75017 Paris, France \\
  \texttt{ey@datakalab.com} \\
}

\usepackage{bibentry}
\begin{document}

\maketitle

\begin{abstract}
Deep neural networks (DNNs) demonstrate outstanding performance across most computer vision tasks. Some critical applications, such as autonomous driving or medical imaging, also require investigation into their behavior and the reasons behind the decisions they make. In this vein, DNN attribution consists in studying the relationship between the predictions of a DNN and its inputs. Attribution methods have been adapted to highlight the most relevant weights or neurons in a DNN, allowing to more efficiently select which weights or neurons can be pruned. However, a limitation of these approaches is that weights are typically compared within each layer separately, while some layers might appear as more critical than others. In this work, we propose to investigate DNN layer importance, \textit{i.e.} to estimate the sensitivity of the accuracy w.r.t. perturbations applied at the layer level. To do so, we propose a novel dataset\footnote{The database and loading scripts are publicly available on github
% \href{https://github.com/publicanonymoussubmission/LayersRanking}{github}
.} to evaluate our method as well as future works. We benchmark a number of criteria and draw conclusions regarding how to assess DNN layer importance and, consequently, how to budgetize layers for increased DNN efficiency (with applications for DNN pruning and quantization), as well as robustness to hardware failure (e.g. bit swaps).
\end{abstract}

\section{Introduction}
Empirical evidence shows the remarkable predictive capabilities of deep neural networks (DNNs). For instance, in computer vision, from image classification \cite{he2016deep} to object detection \cite{he2017mask} through semantic segmentation \cite{chen2017deeplab}, deep neural networks achieve state-of-the-art performance. For a number of applications such as medical imaging \cite{suzuki2017overview} or autonomous driving \cite{grigorescu2020survey}, however, being able to closely understand and monitor the internal behavior of DNNs is of paramount importance. Broadly speaking, this has been related in the literature as DNN explainability \cite{burkart2021survey}.

First, at train time, explainability encompasses the theoretical study of the learning dynamics and the generalization capacities of DNNs \cite{joy1999global,allen2019convergence,zhou2021local}. Second, during deployment, explainability also implies diagnosing \textit{why} a DNN took a particular decision, predicted one class rather than another, or generated one particular sequence of words, conditionally to its inputs. The study of this input-output relationship is often called visual explanation in the context of computer vision, or, more generally, attribution \cite{selvaraju2017grad,shrikumar2017learning,sundararajan2017axiomatic,smilkov2017smoothgrad,adebayo2018sanity,chattopadhay2018grad,novello2022making}. 

Perhaps the most straightforward way to compute and understand attribution for one pixel of an image \cite{zeiler2014visualizing} is to set its value to 0 and measure the accuracy loss induced by this change. However, beyond highlighting the most relevant \textit{pixels} or dimension of an input for prediction, attribution methods can be used to diagnose which \textit{weights} can-or can not-be removed in a DNN for pruning \cite{yvinec2022singe}. One limitation of such approach, however, lies in the fact that weights or neurons can be compared within the same layer only.

In the continuity of this work, we propose to study the cross-layer relative importance with respect to the final accuracy of the model. Having effective methods for layer importance ranking for deep neural networks should open new perspectives on neural networks predictive abilities. Stemming from attribution techniques for input sensitivity understanding and pruning for neuron level sensitivity measurement, we study the effectiveness of these techniques to tackle the challenge of cross-layer sensitivity assessment.

This study paves the way for exciting future perspectives for DNN optimization including, but not limited to:

\begin{itemize}
    \item \textbf{DNN compression (quantization and pruning):} On the one hand, prior work on neural network compression highlighted the necessity to adapt the compression rates to specific layers. For instance, in quantization where floating point operations are converted into low-bit fixed point operations, prior works \cite{krishnamoorthi2018quantizing,nagel2019data,nagel2020up} have shown that quantizing the first and last layers to larger bit-width lead to low latency cost while baring significant benefits in terms of accuracy preservation. Generally speaking, the search for a well suited bit-width assignment per-layer or per neuron is called mixed-precision \cite{wang2019haq,dong2019hawq, cai2020rethinking,prasad2023practical}. However, in practice, it is often either a very costly process based upon a set of simple heuristics \cite{dong2019hawq,prasad2023practical} or reinforcement learning \cite{wang2019haq,cai2020rethinking}. Similarly, in the context of DNN pruning, where one seeks to remove blocks of computations in order to reduce both the inference runtime and the memory footprint of the model, previous work have employed simple strategies to assign a pruning rate per-layer \cite{molchanov2019importance,blalock2020state,yvinec2021red}. These empirical results suggest that the correct assignment of the compression rates is an unsolved important aspect of neural network inference acceleration. 
    \item \textbf{Robust Inference:} On the other hand, multiple critical applications \cite{suzuki2017overview,grigorescu2020survey} of deep neural networks require strong guarantees on the robustness of the predictive function w.r.t. random bit-flips caused by hardware failures. These attacks can give rise to incorrect predictions and lead to system failures that can be catastrophic in the case of critical systems. For example, as pinpointed in \cite{lipp2020nethammer}, it is possible to attack an hardware and induce one bit swap every 350ms on a stream of 500Mbit/s during data transfers in memory. Similarly, on DDR2 memory, hardware manufacturers have measured that on average 22696 errors occur every year. The baseline strategy \cite{liss1986software} to detect these failures and discard the corresponding computations consists in performing said computations twice and ensuring that the results are identical. However, with the growing size of neural networks \cite{zhang2022opt}, such solution could lead to unsustainable energy consumption. Therefore, from a robustness standpoint, there is a growing need to not only be able to compare DNNs neurons or weights \textit{inside} a specific layer, but also to compare the layers \textit{themselves} to only induce redundant calculations to the most important parts of the networks.
\end{itemize}

In what follows, we investigate the challenge of layer-wise sensitivity assessment through the lens of attribution methods. Specifically, we adapt and benchmark a number of attribution methods for fine-grained weight relevance estimation, as well as a number of reduction methods to derive layer-wise importance criteria. In order to evaluate these criteria, we then implement two tests. First, we constructed a dataset of DNN models and their corresponding layers ranking obtained \textit{via} an exhaustive search. Second, we apply these methods in a straightforward manner to DNN compression and robustness to hardware failures in realistic scenarios. Our empirical results enabled us to draw conclusions on best practices regarding cross-layer sensitivity measurement.

%Efficiently ranking layers in terms of importance would enable the design of stronger and simpler mixed precision compression schemes. A second application is robust AI \cite{dietterich2017steps}, in particular, for robustness to hardware failures \cite{muscettola1998remote}. For instance, by duplicating the most important layers and ensuring that the two distinct layers provided identical outputs in order to detect possible bit swaps \cite{lipp2020nethammer}. Such robustness is of paramount importance for safety-critical systems.

\section{Layer-wise sensitivity assessment}\label{methodo}

Let's consider a trained neural network $f$ and each layers $f_l$ with weights $w_l$. In this study, we seek to rank the importance of the layers $(f_1,...,f_L)$ of $f$ using only computations of $f$ on a small, unlabelled, calibration set $\mathcal{X}=(x_1,...,x_n)$ (with $n=256$ in practice), akin to \cite{nagel2020up}. Let $\epsilon_l$ denote a perturbation applied to each weight (or alternatively, to its activations) of layer $f_l$, such that $f^{\epsilon_l}$ is the function disturbed at layer $l$ and $f^{\epsilon_l}_l$ its $l$-th layer. With this convention, we define the importance of layer $l$ as:
\begin{equation}\label{eqdefsens}
    \mathcal{I}(f,\epsilon_l) \triangleq - \mathbb{E}_{X_{\text{test}}}\left[ ||f^{\epsilon_l} - f|| \right]
\end{equation}
where $X_{\text{test}}$ is a test set. Intuitively, if a specific perturbation $\epsilon_l$ applied to layer $l$ causes large changes in the predictive function (as compared to applying the same kind of perturbation to other layers), then this layer is likely to be particularly important. Consequently, our goal is to find a criterion $C$ which predicts the importance ranking of any layer $l$ with respect to a perturbation $\epsilon_l$ of said layer, \textit{i.e.}:
\begin{equation}\label{eq:goal}
    \forall i,j \quad {C(f, \mathcal{X})}_i \leq {C(f, \mathcal{X})}_j \Leftrightarrow \mathcal{I}(f,\epsilon_i) \leq \mathcal{I}(f,\epsilon_j)
\end{equation}
Simply put, computing $C$ upon $\mathcal{X}$ for each layer of $f$ is sufficient to assess the importance of the layers w.r.t. a considered perturbation. Moreso, we are particularly interested in finding importance criteria $C$ that assess layer-wise sensitivity in a general sense, and hence that do not depend on the nature of the perturbations $\epsilon_l$. Such criteria can thus be calculated solely by evaluating statistics of $f$ on the calibration set $\mathcal{X}$. Furthermore, we propose to search for criteria that can be written as:

\begin{equation}\label{eq:2comp}
\centering
C(f, \mathcal{X})_l = \psi \circ \left( \phi(f, \mathcal{X})_l \right)
\end{equation}

Where $\phi$ denotes a function that extracts fine-grained sensitivity information (\textit{i.e.}, at the level of a layer's weight), and $\psi$ reduces this information to an ordered set such as $\mathbb{R}$ to derive a ranking for the layers. Below we first describe existing candidates for function $\phi$.

\subsection{Fine-grained weight relevance estimation}

Authors in \cite{yvinec2022singe} showed that assessing the relevance of a predictive function w.r.t. its weights is a problem that bears similarity with attribution techniques. Inspired by this, we adapt and benchmark several such techniques to design candidates for the weight relevance estimation function $\phi$. Existing candidates include zero and first order criteria that use the weight and gradient values, the higher order methods, as well as the methods derived from the integrated gradients method \cite{sundararajan2017axiomatic}, and lastly, recent black box techniques.

\subsubsection{Zero and First Order Criteria}
\paragraph{Weights:} measuring the weights of a neural network in order to estimate their contribution to the predictive function has been widely studied in pruning \cite{molchanov2019importance}. The resulting function, denoted $W$, offers the advantage of being fairly simple and can be computed without data. However, it does not account for inter-layer relationships.
\begin{equation}
    W : (f, \mathcal{X}) \rightarrow (w_1,...,w_L)
\end{equation}
\paragraph{Gradients:} in GradCam \cite{selvaraju2016grad}, attribution is computed as the gradients of function $f$ w.r.t. each pixel of the image or feature map. This can be adapted by computing the gradients of $f$ w.r.t. each \textit{weight} instead: 
\begin{equation}
    \nabla : (f, \mathcal{X}) \rightarrow \left( \mathbb{E}_{\mathcal{X}}\left[\frac{\partial f}{\partial w_1}\right], ..., \mathbb{E}_{\mathcal{X}}\left[\frac{\partial f}{\partial w_L}\right] \right)
\end{equation}
\paragraph{Weight $\times$ gradients:} combining these two approaches \cite{shrikumar2016not, chen2019delving} usually leads to a slight improvement in practice.
\begin{equation}
    \text{W} \times \nabla: (f, \mathcal{X}) \rightarrow {\left( \mathbb{E}_{\mathcal{X}}\left[w_l\times\frac{\partial f}{\partial w_l}\right]\right)}_{l \in \{1,...,L\}}
\end{equation}

\subsubsection{Higher Order Criterion}
\paragraph{GradCam++:} we also consider the most widely used high order attribution technique\cite{chattopadhay2018grad} and adapt it to weight values as follows
\begin{equation}
    \text{GCam++} \!: (f, \mathcal{X}) \!\rightarrow\!\! {\left( \!\!\mathbb{E}_{\mathcal{X}}\!\!\!\left[\!\frac{{\left(\frac{\partial f}{\partial w_l}\right)}^2}{2 {\left(\frac{\partial f}{\partial w_l}\right)}^2 + w_l{\left(\frac{\partial f}{\partial w_l}\right)}^3}\!\right]\!\right)}_{\!\!\!l \in \{1,...,L\}}
\end{equation}

\subsubsection{Integrated Gradients Criteria}

A known pitfall \cite{yvinec2022singe} of the previously mentioned gradient methods is that the measured importance is by definition local and do not hold when considering important modification of the weight values (such as, for instance, bringing this weight to $0$).

\paragraph{Integrated gradients (IG):} To address this, authors in \cite{sundararajan2017axiomatic,yvinec2022singe} propose to measure the gradients for several values of a considered input (or weight, in our case) on a path towards 0 (\textit{i.e.} for a weight matrix $w_l$ at layer $l$, we consider $\lambda . w_l$ with $\lambda \in [0,1]$). This so-called integrated gradients criterion can be written as:
\begin{equation}
    \text{IG} : (f, \mathcal{X}) \rightarrow {\left( \mathbb{E}_{\mathcal{X}}\left[\sum_{\lambda \in[0;1]}\frac{\partial f}{\partial \lambda w_l}\right]\right)}_{l \in \{1,...,L\}}
\end{equation}
\paragraph{Guided integrated gradients (GIG)} \cite{kapishnikov2021guided} is a refinement of the IG criterion that consists in shrinking, for each integrated gradient iteration, only the least important values, as defined by their gradient magnitudes $||\frac{\partial f}{\partial \lambda w_l}||$.

%following the success of the integrated gradients method \cite{sundararajan2017axiomatic}, many methods were proposed to refine the resulting attribution scores. For instance, GIG  proposes to change the way the parameter $\lambda$ is applied. Instead of being shared (identical convergence to zero for every element to explain) we apply $\lambda$ only to the least important values.
%\begin{equation}
    %\text{GIG} : (f, \mathcal{X}) \rightarrow \left( \mathbb{E}_{\mathcal{X}}\left[\sum_{\lambda \in[0;1]}\frac{\partial f}{\partial \text{gig}(\lambda, w_1)}\right], ..., \mathbb{E}_{\mathcal{X}}\left[\sum_{\lambda \in[0;1]}\frac{\partial f}{\partial \text{gig}(\lambda, w_L)}\right] \right)
%\end{equation}
\paragraph{Important direction guided integrated gradients (IDGI)} \cite{yang2023idgi} is another recent improvement over the IG method, that consists in using the direction of the gradients, weighted by the  difference between the outputs at each integrated gradients iteration.

%consisted in using only the direction given by the gradients normalized by the outputs difference between each update by $\lambda$.
%\begin{equation}
    %\text{IDGI} : (f, \mathcal{X}) \rightarrow \left( \mathbb{E}_{\mathcal{X}}\left[\sum_{\lambda \in[0;1]}\frac{\partial f}{\partial \text{idgi}(\lambda, w_1)}\right], ..., \mathbb{E}_{\mathcal{X}}\left[\sum_{\lambda \in[0;1]}\frac{\partial f}{\partial \text{idgi}(\lambda, w_L)}\right] \right)
%\end{equation}

\subsubsection{Statistical Criteria}

Statistical approaches improve first order criteria by estimating the sensitivity of $f$ within the neighborhood of the weights, as defined by a small additive random noise $\mathcal{N}$.

\paragraph{SmoothGrad}  \cite{smilkov2017smoothgrad} criterion, on the one hand, consists in computing the expected value of the gradient magnitude within this neighborhood:
\begin{equation}
    \text{Smooth}\nabla : (f, \mathcal{X}) \rightarrow {\left( \mathbb{E}_{X_{\text{test}}, \mathcal{N}}\left[\frac{\partial f}{\partial w_l + \mathcal{N}}\right]\right)}_{l \in \{1,...,L\}}
\end{equation}
\paragraph{VarGrad} \cite{adebayo2018sanity} criterion, on the other hand, measures the variance rather than the expectation over the weights neighborhood:
\begin{equation}
    \text{Var}\nabla : (f, \mathcal{X}) \rightarrow {\left( \mathbb{V}_{X_{\text{test}}, \mathcal{N}}\left[\frac{\partial f}{\partial w_l + \mathcal{N}}\right]\right)}_{l \in \{1,...,L\}}
\end{equation}

\subsubsection{Black Box Criterion}
Originally, black box attribution methods aimed at iteratively explaining the sensitivity of a DNN without accessing intermediate activations and weights, hence not using gradients, simply by zeroing out pixels and observing the induced accuracy drop. These methods have however fallen out of flavor due to high computational costs.

\paragraph{HSIC} \cite{novello2022making} is a recent attempt at designing faster black-box attribution methods, that consists in modelling the dependencies between images regions or patches and variations of the predictive function. We propose to adapt this method by grouping together weights belonging to different neurons, and denote this operation $\text{hsic}$:
\begin{equation}
    \text{HSIC} : (f, \mathcal{X}) \rightarrow \left( \mathbb{E}_{\mathcal{X}}\left[\text{hsic}(w_1)\right], ..., \mathbb{E}_{\mathcal{X}}\left[\text{hsic}(w_L)\right] \right)
\end{equation}

\subsection{Reduction Methods}

In addition to the choice of a criterion for fine-grained weight relevance estimation, we also need to propose candidates for the reduction function $\psi$ in Equation \ref{eq:2comp}. This function needs to be a projection to an ordered set such as $\mathbb{R}$. In our experiments, we studied several reduction options, the first and simplest of which is simply the average of the absolute values for each dimension of the computed fine-grained criterion. For instance, if we choose $W$ as function $\phi$, $C(f, \mathcal{X})_l$ boils down to computing the mean absolute value among weights for each layer $l$. Similarly, we investigate using percentiles as well as $l_1$, $l_2$ and $l_\infty$ norms. 

\section{Experiments}
Our empirical evaluation of the proposed criteria is three-fold. First, we propose a novel testbed for evaluating the relevance of each criterion to measure the layer-wise sensitivity, and rank the layers accordingly (equation \ref{eq:goal}). Second, we demonstrate the importance of layer-wise sensitivity assessment for designing stronger baselines for DNN compression (pruning and quantization) and robustness to hardware failures in realistic scenarios.

\subsection{Layer Importance Ranking}\label{layerranksec}

\paragraph{Synthetic dataset:} to evaluate the criteria proposed in Section \ref{methodo} for layer-wise sensitivity assessment, we constructed a dataset of models and their corresponding layers ranking ground truth. Specifically, we considered a simple binary classification task on Moon dataset \cite{pedregosa2011scikit}. We consider various DNN designs, including vanilla sequential networks (similar to VGG \cite{simonyan2014very} architecture), networks that include skip-connection (\textit{skip}) blocks (such as ResNets network family \cite{he2016deep}, with and without stochastic depth (\textit{skip+SD}) \cite{huang2016deep}) as well as transformer (\textit{transfo}) architectures. For each of these architectural designs, we randomly sample the number of blocks or layers (uniformly between 2 and 6) as well as the layer width (uniformly between 8 and 128) for each layer. We trained each network using ADAM optimizer with learning rate $0.01$ for 6 epochs. Every network reached approximately $100\%$ test accuracy.

\paragraph{Layer importance ground truth generation: } to generate the ground truth layer ranking, we apply a perturbation to the weights or activations and measure its impact on the final accuracy, \textit{i.e.} we directly measure $\mathcal{I}(f,\epsilon_l)$ (Equation \ref{eqdefsens}) induced by specific layer-wise perturbations $\epsilon_l$. These perturbations were sampled from different distribution that simulate several scenarios. For each noise distribution, we varied the signal-to-noise ratio in order to evaluate the behavior of the model w.r.t. more or less difficult settings. 

\begin{itemize}
    \item Multiplicative impulse (pepper), denoted $\mathcal{U}$ where a uniformly drawn random subset (corresponding to a proportion between 0 and $100\%$) of weights or activations are set to 0. This corresponds to unstructured (\textit{i.e.} at the weight level) or structured (\textit{i.e.} at the channel or neuron level) respectively.
    \item Additive Gaussian noise $\mathcal{N}(0, \sigma)$, with $\sigma \in ]0, max(w \in w_l)]$ applied to either weights or activations. This setting bears similarities with quantization process as small, additive perturbations are added to each weight or activation.
    \item Additive impulse or Dirac $\mathcal{D}$ noise where a large perturbation (between 0 and $max(w \in w_l)$) is applied to a proportion between 0 and $100\%$ of uniformly drawn random weights or activations. This bears similarity with random bit swaps as a restricted set of weights or activations undergoes significant, non-zero changes.
\end{itemize}

In what follows, we evaluate the capacity of each criterion (as a combination of a fine-grained criterion and reduction method, both of which will be assessed separately) to guess the correct order (as indicated by $\mathcal{I}(f,\epsilon_l)$ for all above mentioned perturbations $\epsilon_l$): this setting is however very challenging as each order has to be retrieved exactly. Lastly, we ensured that the dataset was not biased \textit{i.e.} that there is sufficient diversity in term of layer ranking for the different architectures and perturbations. These elements can be found in the dataset online description.

\paragraph{Empirical criteria validation and comparison: }

First, a comparison between the different weight relevance criteria is summarized in Table \ref{tab:toy}. First, on average it appears that zeroth and first order attribution techniques such as gradients ($\nabla$) and weight $\times$ gradient ($W \times \nabla$) achieve the highest results on par with more sophisticated criterion like GradCam++ at lower computational cost. Second, when looking specifically at more complicated DNN architectures such as networks with skip-connections and transformers, we observe that the more naive techniques do not work well while more recent and complex techniques such as GradCam++, SmoothGrad, Vargrad and HSIC achieve the best results. However, no method truly achieves satisfactory results on transformers in this challenging setting.

Second, Table \ref{tab:reduction} shows results for the different reduction methods, averaged among all weight relevance criteria. Using $l_1$ norm achieves the worst results as, contrary to the simple average reduction method, it does not normalize the importance measurement w.r.t. the layer width. Overall the best results are obtained using $l_\infty$ norm, \textit{i.e.} intuitively considering the largest sensitivity (\textit{i.e.} fine-grained weight relevance) across all neurons in the layer.

Generally speaking, There is no all-around best fine-grained weight relevance method that works for all architectures and perturbations. However, we can highlight a few key takeaways:
\begin{itemize}
    \item The $\nabla$, $W \times \nabla$, GradCam++ and IDGI methods are generally speaking solid candidates criteria for weight relevance estimation.
    \item Statistical criteria such as Var$\nabla$ are good choices for residual architectures.
    \item GradCam++, Smooth$\nabla$ and Var$\nabla$ are the best for transformers though not as reliable as for other architectures.
    \item $l_\infty$ norm is the best reduction method in all tested cases.
\end{itemize}

Bearing this in mind, in what follows, we apply layer-wise importance ranking to DNN compression as well as robust inference.

\input{Tables/toy}

\begin{table}[t]
    \centering
    \begin{tabular}{|c|c|}
    \hline
    Reduction Method & Avg Score\\% (ranking) & Avg Score (pruning) & Avg Score (robustness) \\
    \hline
    average & 37.086\\% & 49.311 & 73.740 \\
    best percentile & 38.636\\% & 52.758 & 74.191 \\
    $l_1$ & 36.109\\% & 45.290 & 73.389 \\
    $l_2$ & 38.234\\% & 53.010\\% & 74.058 \\
    $l_\infty$ & \textbf{39.809}\\% & \textbf{63.000} & \textbf{74.518} \\
    \hline
    \end{tabular}
    \caption{Average score (number of correct complete orderings) for each reduction method.}
    \label{tab:reduction}
\end{table}

\begin{figure}[t]
    \centering
    \includegraphics[width = \linewidth]{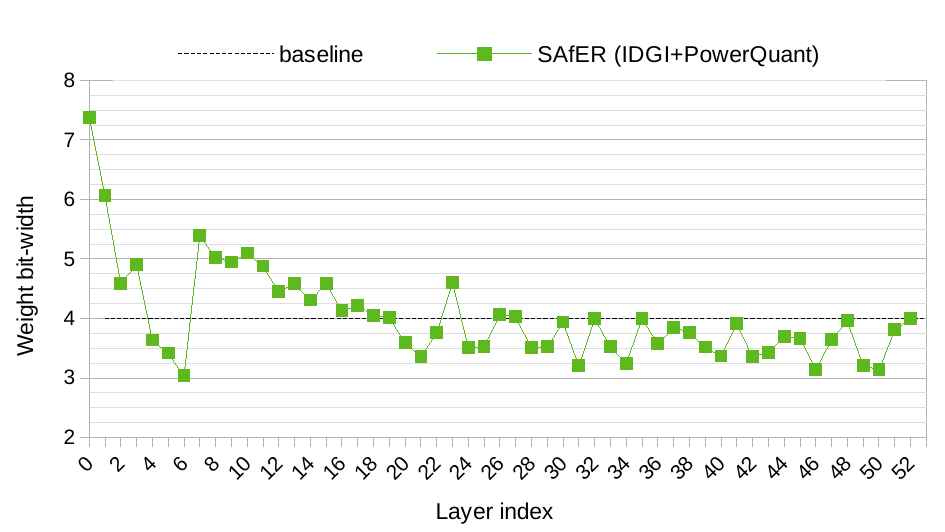}
    \caption{Mixed precision assignment using IDGI \cite{yang2023idgi} and PowerQuant \cite{yvinec2023powerquant}.}
    \label{fig:mixed_precsion}
\end{figure}

\subsection{Applications to DNN Compression}

\subsubsection{Experiments on pruning}

\paragraph{From layer-wise relevance to pruning budgetization:} Given a target pruning rate $\gamma$ for DNN $f$ with weights $w$, we ought to remove $\gamma \sum_l \Omega(w_l)$, where $\Omega(w_l)$ denotes the number of weights in $w_l$. We simply assign the per-layer pruning rates $\gamma_l$ based on the weighting given by the importance score:
\begin{equation}\label{eq:pruning}
    \gamma_l = \alpha * \gamma * {C(f, \mathcal{X})}_l
\end{equation}
with $\alpha$ a normalizing constant such that $\gamma \sum_l \#w_l =  \sum_l \gamma_l \#w_l$.
    
\input{Tables/pruning}

\paragraph{DNN pruning results:} in the frame of TinyML perf challenge \cite{tinymlperf} we use the proposed criteria to set the budget and prune a ResNet-8 network, removing $20\%$ of neurons (\textit{i.e.} structured pruning). Table \ref{tab:pruning} shows the accuracies of the models pruned using all the criteria introduced in Section \ref{methodo} for layer-wise relevance estimation and budgeting (in rows, with $l_\infty$ as the reduction method) and the same criteria for selecting neurons to prune within each layer (in columns). First, overall, we observe a wide discrepancy in the accuracies averaged among columns (last line) as well as among rows (last column): this suggests that, while the method for intra-layer neuron pruning, onto which the community has been focusing so far,  is very important, the inter-layer relevance and budgetization is also of paramount importance. Second, we observe that $\nabla$ and $Var\nabla$ performs the best overall for layer-wise relevance in that context. In particular, $\nabla$ for layer-wise relevance assessment combined with IDGI as neuron pruning criterion offers the best performance, with GradCam++, $Smooth\nabla$ and $Var\nabla$ also working well in tandem with integrated gradient (IG, GIG, IDGI) neuron selection criteria, echoing the results from \cite{yvinec2022singe}. This confirms the results obtained in Section \ref{layerranksec}. Overall, these results motivate further research and distinction between inter and intra-layer importance evaluation criteria.

\subsubsection{Experiments on DNN quantization}

\input{Tables/quantization}

\paragraph{From layer-wise relevance to budgeting layers for quantization:} given a ranking of layers $f_1,...,f_L$ we assign a layer-wise quantization bit-width based on a target average bit-width $b$. For instance, if $b=4$, we assign $3$ bits to the least important third of the layers, $5$ bits to the most important third and $4$ to the remaining layers. Furthermore, as in \cite{nagel2019data} we quantize activations to 8 bits. Also, contrary to prior work \cite{ni2020wrapnet,nagel2019data}, we do not apply any arbitrary quantization bit-width to the first or last layer as our goal is to show the ability of the proposed criteria to properly rank layers without manual intervention or ad hoc heuristics.

\paragraph{Quantization results:} We implement the aforementioned mixed precision scheme (using the proposed criteria to set the bit-width budget for each layer) on a ResNet-50 pretrained on ImageNet and compare with a baseline constant precision quantization in four settings: W6/A8 using two state-of-the-art methods DFQ \cite{nagel2019data} and SQuant \cite{squant2022}, as well as W4/A4 using PowerQuant \cite{yvinec2023powerquant} and Adaround \cite{nagel2020up}. Table \ref{tab:quantization} shows the test accuracies of the quantized networks. Once again, we observe significant discrepancies between the different criteria, with e.g. $W$ providing results significantly below the baseline and $\nabla$ and Var$\nabla$ systematically improving over it. This shows that estimating layer relevance is crucial to the performance of mixed precision quantization. Overall, as in previous experiments, $\nabla$, $W \times \nabla$, GradCam++ as well as Var$\nabla$ perform well on this benchmark, with the addition of IDGI with PowerQuant. 

Figure \ref{fig:mixed_precsion} illustrates the budget obtained using using IDGI \cite{yang2023idgi} as layer-wise relevance criterion over PowerQuant \cite{yvinec2023powerquant} as the baseline quantization method in W4/A8. What's interesting is that our approach sets a high budget to the first layer of the network (as well as a decreasing bit-width budget trend towards the end of the network): this confirms the importance of assigning a larger bit-width to the first layer of the model, as already empirically remarked in \cite{nagel2019data,nagel2020up}.

% which can be attributed to the use of a non-uniform quantization scheme which likely better fits with a less local criterion for layer-wise.

To sum it up, these results on DNN pruning and quantization suggest that assessing layer-wise relevance and using it to budgetize layers in a simple, straightforward way is already enough to improve existing pruning and quantization techniques, which is remarkable considering that implementing successful post-training, few-shot mixed precision schemes is non-trivial in practice \cite{prasad2023practical}. In what follows, we show that layer-wise ranking also find applications for robust inference.

\subsection{Robustness to Hardware Failure}

\paragraph{Layer ranking and robust inference:} in this section, we consider the problem of ensuring robustness w.r.t. random bit-swaps caused by hardware failures that can occur at inference time e.g. during memory transfers or weight loading. A common solution to overcome this is to verify the computations performed by a layer, \textit{i.e.} by performing these computations twice and comparing the results \footnote{We share an implementation
% \href{https://github.com/publicanonymoussubmission/bitswapdetection}{implementation}
in both Torch and TensorFlow to facilitate further research using this evaluation.}. To limit the computation overhead induced by redundant computations, we verify only the most important layers, as budgeted by one of the aforementioned criteria. Intuitively, if we rank the layers by importance and gradually increase the number of layers left unchecked under random bit swaps, the accuracy will decrease: in such a case, the slower it decreases, the better the ranking criterion for robust inference. 

\begin{figure}
\begin{center}
\includegraphics[width = \linewidth]{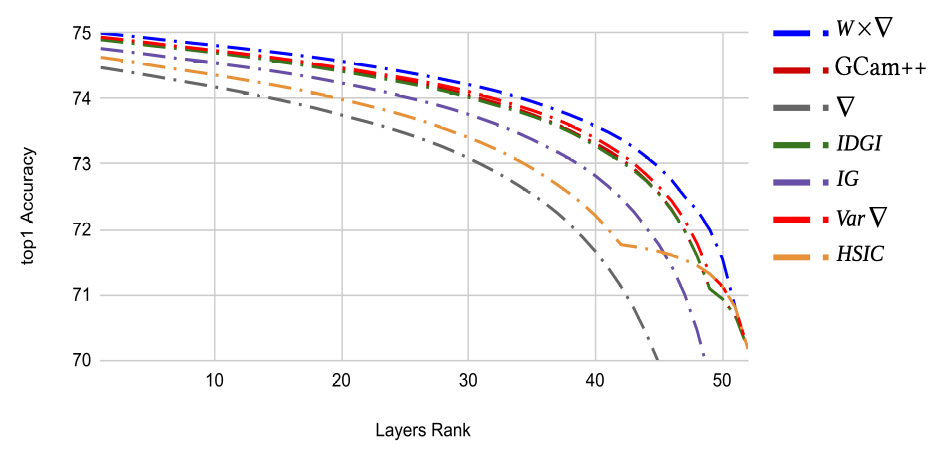}
\vspace{-0.5cm}
\caption{Evolution of the accuracy of a ResNet 50 trained on ImageNet as we progressively left important layers unchecked for random bit-swaps. All curves trend downwards as less bit-swap are detected hence increasing the damage to the predictive function. The higher the accuracy, the better.}
\label{fig:robustnessimagenet}
\end{center}
\end{figure}

\paragraph{Robust inference results:} In Figure \ref{fig:robustnessimagenet}, we report the evolution of the accuracy of the model under random bit-swaps with respect to the number of layers that remain unchecked (starting from 1). Similarly to what precedes, we observe the good performance of the $W \times \nabla$ and GradCam++ criteria. However, in this case, $\nabla$ is the least performing criterion, which can be attributed to the fact that gradients only measure local changes and specifically target the weights, while the bit-swaps are randomly applied to both weights and activations and can induce huge changes that break this locality principle. Conversely, other methods such as $W \times \nabla$, GradCam++ and IDGI explicitly use information on both weights and gradients for a more relevant sensitivity assessment in this context. Nonetheless, we show that it is possible to preserve the accuracy without more that 1\% accuracy loss while only checking 17 layers out of 52 which significantly reduces the redundancy overhead.

\section{Conclusion}

In this work, we study and pave the way for future research on layer-wise sensitivity of Deep Neural Nets. In particular, we stem from DNN attribution techniques for input sensitivity understanding, and adapt these to derive candidates for fine-grained sensitivity assessment of the whole predictive function w.r.t. particular weights in a layer. We also list a number of candidates of reduction methods to integrate these fine-grained information into a layer-wise measurement. In order to evaluate these methods and future works, we designed a synthetic dataset of neural network architectures from sequential to more complex designs with an exhaustive study of the sensitivity of the predictive function w.r.t. perturbations applied to each layer as the ground truth. We experimentally demonstrated that it is possible to retrieve the correct layer ranking in this setting, as well as derive best practices for layer-wise sensitivity assessment. We then applied this framework for two practical applications: First, DNN compression \textit{via} pruning and quantization (mixed precision). In this setup, we show that, with little effort, we can improve the performance of these methods by straightforwardly translating cross-layer relevance measurements into budgets for compression. Second, for robust inference, we can apply our layer-wise sensitivity assessment to check only the most relevant layers and avoid random bit-swaps caused by hardware failures.

\paragraph{Limitations and future Work:} First, one limitation of the proposed work is that the all the proposed candidates criteria fail to achieve satisfactory performance on transformers which are currently taking over NLP and computer vision domains. Perhaps the use of more complex and specific criteria shall be considered to solve this issue. Second, the ideas proposed in this paper could offer strong benefits towards more efficient DNN design: for instance, balancing per-layer importance and runtime cost (e.g. in a framework similar to \cite{shen2022structural}) could lead to more practical architectures, as well as pruning and quantization schemes. Furthermore, the study of intermediate levels of granularity (e.g. neurons, group of neurons or computational blocks) could lead to even more efficient inference as well as less costly monitoring of hardware failures. 

\textbf{Acknowledgments:} This work has been supported by the french National Association for Research and Technology (ANRT), the company Datakalab (CIFRE convention C20/1396) and by the French National Agency (ANR)  (FacIL, project ANR-17-CE33-0002). This work was granted access to the HPC resources of IDRIS under the allocation 2022-AD011013384 made by GENCI. This work has been supported by the French government under the "France 2030” program, as part of the SystemX Technological Research Institute within the Confiance.ai project.

\bibliography{aaai24}

\end{document}

%% file: Tables/toy.tex
\begin{table*}[!t]
\centering
\caption{For each family of architectures (Vanilla, skip, skip+SD and transfo) and distribution prior ($\mathcal{U}$, Gaussian $\mathcal{N}$ and $\mathcal{D}$, applied to weights (W) or activations (act)), we report the proportion of correct rankings of each criterion (with $l_\infty$ as the reduction method). The scores are averaged over a $1000$ sampled architectures per row.}
\label{tab:toy}
\begin{tabular}{cccccccccccc}
\hline
\multicolumn{1}{|c|}{Archi} & \multicolumn{1}{|l|}{Noise} & \multicolumn{1}{c|}{W} & \multicolumn{1}{c|}{$\nabla$} & \multicolumn{1}{c|}{$\text{W} \times \nabla$} & \multicolumn{1}{c|}{\cellcolor[HTML]{C0C0C0}GCam++} & \multicolumn{1}{c|}{IG} & \multicolumn{1}{c|}{GIG}  & \multicolumn{1}{c|}{IDGI} & \multicolumn{1}{c|}{Smooth$\nabla$} & \multicolumn{1}{c|}{Var$\nabla$} & \multicolumn{1}{c|}{Hsic} \\ \hline
\multicolumn{1}{|c|}{Vanilla} &\multicolumn{1}{|c|}{$\mathcal{U}$ on W} & \multicolumn{1}{c|}{25} & \multicolumn{1}{c|}{76}&  \multicolumn{1}{c|}{76} & \multicolumn{1}{c|}{\cellcolor[HTML]{C0C0C0}76} & \multicolumn{1}{c|}{59} & \multicolumn{1}{c|}{49} & \multicolumn{1}{c|}{71} & \multicolumn{1}{c|}{71} & \multicolumn{1}{c|}{28} & \multicolumn{1}{c|}{38} \\
\multicolumn{1}{|c|}{Vanilla} &\multicolumn{1}{|c|}{$\mathcal{N}$ on W} & \multicolumn{1}{c|}{42} & \multicolumn{1}{c|}{68}&  \multicolumn{1}{c|}{68} & \multicolumn{1}{c|}{\cellcolor[HTML]{C0C0C0}68} & \multicolumn{1}{c|}{59} & \multicolumn{1}{c|}{56} & \multicolumn{1}{c|}{66} & \multicolumn{1}{c|}{66} & \multicolumn{1}{c|}{36} & \multicolumn{1}{c|}{47} \\
\multicolumn{1}{|c|}{Vanilla} &\multicolumn{1}{|c|}{$\mathcal{D}$ on W} & \multicolumn{1}{c|}{32} & \multicolumn{1}{c|}{60}&  \multicolumn{1}{c|}{60} & \multicolumn{1}{c|}{\cellcolor[HTML]{C0C0C0}33} & \multicolumn{1}{c|}{41} & \multicolumn{1}{c|}{60} & \multicolumn{1}{c|}{48} & \multicolumn{1}{c|}{60} & \multicolumn{1}{c|}{43} & \multicolumn{1}{c|}{60} \\ \hline
\multicolumn{1}{|c|}{Vanilla} &\multicolumn{1}{|c|}{$\mathcal{U}$ on act} & \multicolumn{1}{c|}{82} & \multicolumn{1}{c|}{82}&  \multicolumn{1}{c|}{82} & \multicolumn{1}{c|}{\cellcolor[HTML]{C0C0C0}33} & \multicolumn{1}{c|}{67} & \multicolumn{1}{c|}{78} & \multicolumn{1}{c|}{78} & \multicolumn{1}{c|}{53} & \multicolumn{1}{c|}{29} & \multicolumn{1}{c|}{77} \\
\multicolumn{1}{|c|}{Vanilla} &\multicolumn{1}{|c|}{$\mathcal{N}$ on act} & \multicolumn{1}{c|}{82} & \multicolumn{1}{c|}{82}&  \multicolumn{1}{c|}{82} & \multicolumn{1}{c|}{\cellcolor[HTML]{C0C0C0}33} & \multicolumn{1}{c|}{67} & \multicolumn{1}{c|}{78} & \multicolumn{1}{c|}{78} & \multicolumn{1}{c|}{53} & \multicolumn{1}{c|}{29} & \multicolumn{1}{c|}{63} \\
\multicolumn{1}{|c|}{Vanilla} &\multicolumn{1}{|c|}{$\mathcal{D}$ on act} & \multicolumn{1}{c|}{82} & \multicolumn{1}{c|}{82}&  \multicolumn{1}{c|}{82} & \multicolumn{1}{c|}{\cellcolor[HTML]{C0C0C0}33} & \multicolumn{1}{c|}{67} & \multicolumn{1}{c|}{77} & \multicolumn{1}{c|}{77} & \multicolumn{1}{c|}{55} & \multicolumn{1}{c|}{28} & \multicolumn{1}{c|}{49} \\ \hline\hline
\multicolumn{1}{|c|}{skip} & \multicolumn{1}{|c|}{$\mathcal{U}$ on W} & \multicolumn{1}{c|}{46} & \multicolumn{1}{c|}{47}&  \multicolumn{1}{c|}{43} & \multicolumn{1}{c|}{\cellcolor[HTML]{C0C0C0}64} & \multicolumn{1}{c|}{44} & \multicolumn{1}{c|}{43} & \multicolumn{1}{c|}{47} & \multicolumn{1}{c|}{52} & \multicolumn{1}{c|}{77} & \multicolumn{1}{c|}{28} \\
\multicolumn{1}{|c|}{skip} & \multicolumn{1}{|c|}{$\mathcal{N}$ on W} & \multicolumn{1}{c|}{23} & \multicolumn{1}{c|}{31}&  \multicolumn{1}{c|}{32} & \multicolumn{1}{c|}{\cellcolor[HTML]{C0C0C0}71} & \multicolumn{1}{c|}{27} & \multicolumn{1}{c|}{33} & \multicolumn{1}{c|}{28} & \multicolumn{1}{c|}{28} & \multicolumn{1}{c|}{39} & \multicolumn{1}{c|}{34} \\
\multicolumn{1}{|c|}{skip} & \multicolumn{1}{|c|}{$\mathcal{D}$ on W} & \multicolumn{1}{c|}{30} & \multicolumn{1}{c|}{30}&  \multicolumn{1}{c|}{32} & \multicolumn{1}{c|}{\cellcolor[HTML]{C0C0C0}72} & \multicolumn{1}{c|}{28} & \multicolumn{1}{c|}{31} & \multicolumn{1}{c|}{27} & \multicolumn{1}{c|}{31} & \multicolumn{1}{c|}{47} & \multicolumn{1}{c|}{16} \\ \hline\hline
\multicolumn{1}{|c|}{skip+SD} & \multicolumn{1}{|c|}{$\mathcal{U}$ on W} & \multicolumn{1}{c|}{42} & \multicolumn{1}{c|}{42}&  \multicolumn{1}{c|}{42} & \multicolumn{1}{c|}{\cellcolor[HTML]{C0C0C0}25} & \multicolumn{1}{c|}{41} & \multicolumn{1}{c|}{42} & \multicolumn{1}{c|}{42} & \multicolumn{1}{c|}{17} & \multicolumn{1}{c|}{10} & \multicolumn{1}{c|}{22} \\
\multicolumn{1}{|c|}{skip+SD} & \multicolumn{1}{|c|}{$\mathcal{N}$ on W} & \multicolumn{1}{c|}{21} & \multicolumn{1}{c|}{24}&  \multicolumn{1}{c|}{24} & \multicolumn{1}{c|}{\cellcolor[HTML]{C0C0C0}53} & \multicolumn{1}{c|}{23} & \multicolumn{1}{c|}{23} & \multicolumn{1}{c|}{22} & \multicolumn{1}{c|}{22} & \multicolumn{1}{c|}{32} & \multicolumn{1}{c|}{40} \\
\multicolumn{1}{|c|}{skip+SD} & \multicolumn{1}{|c|}{$\mathcal{D}$ on W} & \multicolumn{1}{c|}{20} & \multicolumn{1}{c|}{21}&  \multicolumn{1}{c|}{17} & \multicolumn{1}{c|}{\cellcolor[HTML]{C0C0C0}47} & \multicolumn{1}{c|}{19} & \multicolumn{1}{c|}{19} & \multicolumn{1}{c|}{20} & \multicolumn{1}{c|}{20} & \multicolumn{1}{c|}{29} & \multicolumn{1}{c|}{12} \\ \hline\hline
\multicolumn{1}{|c|}{transfo} & \multicolumn{1}{|c|}{$\mathcal{U}$ on W} & \multicolumn{1}{c|}{0} & \multicolumn{1}{c|}{0} & \multicolumn{1}{c|}{0} & \multicolumn{1}{c|}{\cellcolor[HTML]{C0C0C0}22} & \multicolumn{1}{c|}{0} & \multicolumn{1}{c|}{0} & \multicolumn{1}{c|}{0} & \multicolumn{1}{c|}{6} & \multicolumn{1}{c|}{12} &  \multicolumn{1}{c|}{0} \\
\multicolumn{1}{|c|}{transfo} & \multicolumn{1}{|c|}{$\mathcal{N}$ on W} & \multicolumn{1}{c|}{1} & \multicolumn{1}{c|}{1} & \multicolumn{1}{c|}{1} & \multicolumn{1}{c|}{\cellcolor[HTML]{C0C0C0}10} & \multicolumn{1}{c|}{0} & \multicolumn{1}{c|}{1} & \multicolumn{1}{c|}{1} & \multicolumn{1}{c|}{18} & \multicolumn{1}{c|}{12} &  \multicolumn{1}{c|}{7} \\
\multicolumn{1}{|c|}{transfo} & \multicolumn{1}{|c|}{$\mathcal{D}$ on W} & \multicolumn{1}{c|}{0} & \multicolumn{1}{c|}{0} & \multicolumn{1}{c|}{0} & \multicolumn{1}{c|}{\cellcolor[HTML]{C0C0C0}7} & \multicolumn{1}{c|}{0} & \multicolumn{1}{c|}{0} & \multicolumn{1}{c|}{0} & \multicolumn{1}{c|}{5} & \multicolumn{1}{c|}{14} &  \multicolumn{1}{c|}{0} \\ \hline
\\ \hline
\multicolumn{2}{|c|}{avg } & \multicolumn{1}{c|}{35} & \multicolumn{1}{c|}{43} & \multicolumn{1}{c|}{43} & \multicolumn{1}{c|}{\cellcolor[HTML]{C0C0C0}43} & \multicolumn{1}{c|}{36} & \multicolumn{1}{c|}{39} & \multicolumn{1}{c|}{40} & \multicolumn{1}{c|}{37} & \multicolumn{1}{c|}{31} & \multicolumn{1}{c|}{33} \\ \hline
\end{tabular}
\end{table*}

%% file: Tables/pruning.tex
\begin{table*}[!t]
\centering
\caption{Results ($\%$acc) obtained for TinyML perf challenge ResNet-8 on Cifar-10 with various inter-layer (rows) and intra-layer (columns) relevance criteria. Accuracies are reported for a global pruning rate of $\gamma=20\%$ of a ResNet-8 trained on Cifar-10 averaged over 10 runs and without post-pruning fine-tuning.}
\label{tab:pruning}
\begin{tabular}{|c|c|c|c|c|c|c|>{\columncolor[HTML]{C0C0C0}}c|c|c|c|c|}
\hline
\multicolumn{1}{|l|}{} & W & $\nabla$ & W $\times \nabla$ & GCam++  & IG & GIG & IDGI & Smooth$\nabla$ & Var$\nabla$       & Hsic & avg \\ \hline
W                 & 31 & 63 & 18 & 41 & 75 & 75 & 73 & 32 & 21 & 32 & 46 \\ \hline
\cellcolor[HTML]{C0C0C0}$\nabla$ & \cellcolor[HTML]{C0C0C0}31 & \cellcolor[HTML]{C0C0C0}71 & \cellcolor[HTML]{C0C0C0}79 & \cellcolor[HTML]{C0C0C0}67 &\cellcolor[HTML]{C0C0C0}80 & \cellcolor[HTML]{C0C0C0}80& 82 & \cellcolor[HTML]{C0C0C0}55 & \cellcolor[HTML]{C0C0C0}13 & \cellcolor[HTML]{C0C0C0}74 & \cellcolor[HTML]{C0C0C0}63 \\ \hline
W $\times \nabla$ & 31 & 41 & 47 & 32 & 75 & 75 & 75 & 37 & 37 & 50 & 50 \\ \hline
GCam++            & 31 & 27 & 82 & 70 & 80 & 80 & 80 & 12 & 52 & 22 & 54 \\ \hline
IG                & 31 & 32 & 64 & 23 & 73 & 75 & 75 & 37 & 65 & 39 & 51 \\ \hline
GIG               & 31 & 32 & 66 & 59 & 75 & 73 & 76 & 32 & 47 & 38 & 53 \\ \hline
IDGI              & 23 & 21 & 55 & 64 & 76 & 73 & 76 & 28 & 50 & 59 & 53 \\ \hline
Smooth$\nabla$    & 31 & 17 & 69 & 33 & 50 & 80 & 80 & 62 & 47 & 65 & 53 \\ \hline
Var$\nabla$       & 31 & 77 & 75 & 75 & 80 & 80 & 80 & 63 & 28 & 19 & 61 \\ \hline
Hsic              & 31 & 36 & 57 & 44 & 73 & 75 & 75 & 11 & 26 & 39 & 47 \\ \hline
avg               & 30 & 42 & 61 & 51 & 74 & 77 & 77 & 33 & 43 & 44 & \multicolumn{1}{l|}{} \\ \hline
\end{tabular}
\end{table*}

%% file: Tables/quantization.tex
\begin{table*}[!t]
\caption{Results ($\%$acc) obtained for mixed-precision quantization of a ResNet-50 on ImageNet, using DFQ \cite{nagel2019data}, SQuant \cite{squant2022} for an average bit-width of $6$ and exactly $8$ (W6/A8) for the weights and activations, respectively, as well as PowerQuant \cite{yvinec2023powerquant} and AdaRound \cite{nagel2020up} for an average of $4$ bits and exactly $8$ bits for the weights and activations.}
\label{tab:quantization}
\centering
\begin{tabular}{ccc|l|ccc|}
\cline{1-3} \cline{5-7}
\multicolumn{3}{|c|}{DFQ - ICCV '19} &  & \multicolumn{3}{c|}{SQuant - ICLR '22} \\ \cline{1-3} \cline{5-7} 
\multicolumn{1}{|c|}{method} & \multicolumn{1}{c|}{quantization} & accuracy &  & \multicolumn{1}{c|}{method} & \multicolumn{1}{c|}{quantization} & accuracy \\ \cline{1-3} \cline{5-7} 
\multicolumn{1}{|c|}{W} & \multicolumn{1}{c|}{\multirow{9}{*}{W4-8/A8}} & 73.282 &  & \multicolumn{1}{c|}{W} & \multicolumn{1}{c|}{\multirow{9}{*}{W4-8/A8}} & 74.462 \\ \cline{1-1} \cline{3-3} \cline{5-5} \cline{7-7} 
\multicolumn{1}{|c|}{$\nabla$} & \multicolumn{1}{c|}{} & 74.636 &  & \multicolumn{1}{c|}{$\nabla$} & \multicolumn{1}{c|}{} & 74.776 \\ \cline{1-1} \cline{3-3} \cline{5-5} \cline{7-7} 
\multicolumn{1}{|c|}{W $\times \nabla$} & \multicolumn{1}{c|}{} & \textit{74.724} &  & \multicolumn{1}{c|}{W $\times \nabla$} & \multicolumn{1}{c|}{} & \textbf{74.864} \\ \cline{1-1} \cline{3-3} \cline{5-5} \cline{7-7} 
\multicolumn{1}{|c|}{GCam++} & \multicolumn{1}{c|}{} & 74.546 &  & \multicolumn{1}{c|}{GCam++} & \multicolumn{1}{c|}{} & 74.706 \\ \cline{1-1} \cline{3-3} \cline{5-5} \cline{7-7} 
\multicolumn{1}{|c|}{IG} & \multicolumn{1}{c|}{} & 74.710 &  & \multicolumn{1}{c|}{IG} & \multicolumn{1}{c|}{} & 74.212 \\ \cline{1-1} \cline{3-3} \cline{5-5} \cline{7-7} 
\multicolumn{1}{|c|}{IDGI} & \multicolumn{1}{c|}{} & \textbf{74.816} &  & \multicolumn{1}{c|}{IDGI} & \multicolumn{1}{c|}{} & 74.496 \\ \cline{1-1} \cline{3-3} \cline{5-5} \cline{7-7} 
\multicolumn{1}{|c|}{Smooth$\nabla$} & \multicolumn{1}{c|}{} & 65.284 &  & \multicolumn{1}{c|}{Smooth$\nabla$} & \multicolumn{1}{c|}{} & 74.616 \\ \cline{1-1} \cline{3-3} \cline{5-5} \cline{7-7} 
\multicolumn{1}{|c|}{Var$\nabla$} & \multicolumn{1}{c|}{} & 74.284 &  & \multicolumn{1}{c|}{Var$\nabla$} & \multicolumn{1}{c|}{} & \textit{74.848} \\ \cline{1-1} \cline{3-3} \cline{5-5} \cline{7-7} 
\multicolumn{1}{|c|}{HSIC} & \multicolumn{1}{c|}{} & 73.902 &  & \multicolumn{1}{c|}{HSIC} & \multicolumn{1}{c|}{} & 74.600 \\ \cline{1-1} \cline{3-3} \cline{5-5} \cline{7-7} 
\cline{1-3} \cline{5-7} 
\multicolumn{1}{|c|}{-} & \multicolumn{1}{c|}{W6/A8} & 73.904 &  & \multicolumn{1}{c|}{-} & \multicolumn{1}{c|}{W6A8} & 74.596 \\ \cline{1-3} \cline{5-7} 
\multicolumn{1}{|c|}{-} & \multicolumn{1}{c|}{full-precision} & 75.000 &  & \multicolumn{1}{c|}{-} & \multicolumn{1}{c|}{full-precision} & 75.000 \\ \cline{1-3} \cline{5-7} 
\\
\cline{1-3} \cline{5-7}
\multicolumn{3}{|c|}{PowerQuant - ICLR '23} &  & \multicolumn{3}{c|}{AdaRound - ICML '20} \\ \cline{1-3} \cline{5-7} 
\multicolumn{1}{|c|}{method} & \multicolumn{1}{c|}{quantization} & accuracy &  & \multicolumn{1}{c|}{method} & \multicolumn{1}{c|}{quantization} & accuracy \\ \cline{1-3} \cline{5-7} 
\multicolumn{1}{|c|}{W} & \multicolumn{1}{c|}{\multirow{9}{*}{W3-5/A8}} & 41.426 &  & \multicolumn{1}{c|}{W} & \multicolumn{1}{c|}{\multirow{9}{*}{W3-5/A8}} & 71.958 \\ \cline{1-1} \cline{3-3} \cline{5-5} \cline{7-7} 
\multicolumn{1}{|c|}{$\nabla$} & \multicolumn{1}{c|}{} & 69.894 &  & \multicolumn{1}{c|}{$\nabla$} & \multicolumn{1}{c|}{} & 74.052 \\ \cline{1-1} \cline{3-3} \cline{5-5} \cline{7-7} 
\multicolumn{1}{|c|}{W $\times\nabla$} & \multicolumn{1}{c|}{} & \textit{70.978} &  & \multicolumn{1}{c|}{W $\times\nabla$} & \multicolumn{1}{c|}{} & \textbf{74.256} \\ \cline{1-1} \cline{3-3} \cline{5-5} \cline{7-7} 
\multicolumn{1}{|c|}{GCam++} & \multicolumn{1}{c|}{} & 70.920 &  & \multicolumn{1}{c|}{GCam++} & \multicolumn{1}{c|}{} & 73.936 \\ \cline{1-1} \cline{3-3} \cline{5-5} \cline{7-7} 
\multicolumn{1}{|c|}{IG} & \multicolumn{1}{c|}{} & 69.168 &  & \multicolumn{1}{c|}{IG} & \multicolumn{1}{c|}{} & 71.766 \\ \cline{1-1} \cline{3-3} \cline{5-5} \cline{7-7} 
\multicolumn{1}{|c|}{IDGI} & \multicolumn{1}{c|}{} & \textbf{73.070} &  & \multicolumn{1}{c|}{IDGI} & \multicolumn{1}{c|}{} & 72.172 \\ \cline{1-1} \cline{3-3} \cline{5-5} \cline{7-7} 
\multicolumn{1}{|c|}{Smooth$\nabla$} & \multicolumn{1}{c|}{} & 52.310 &  & \multicolumn{1}{c|}{Smooth$\nabla$} & \multicolumn{1}{c|}{} & 72.498 \\ \cline{1-1} \cline{3-3} \cline{5-5} \cline{7-7} 
\multicolumn{1}{|c|}{Var$\nabla$} & \multicolumn{1}{c|}{} & 66.722 &  & \multicolumn{1}{c|}{Var$\nabla$} & \multicolumn{1}{c|}{} & \textit{74.246} \\ \cline{1-1} \cline{3-3} \cline{5-5} \cline{7-7} 
\multicolumn{1}{|c|}{HSIC} & \multicolumn{1}{c|}{} & 69.120 &  & \multicolumn{1}{c|}{HSIC} & \multicolumn{1}{c|}{} & 73.936 \\ 
\cline{1-3} \cline{5-7} 
\multicolumn{1}{|c|}{-} & \multicolumn{1}{c|}{W4/A8} & 63.408 &  & \multicolumn{1}{c|}{-} & \multicolumn{1}{c|}{W4A8} & 73.978 \\ \cline{1-3} \cline{5-7} 
\multicolumn{1}{|c|}{-} & \multicolumn{1}{c|}{full-precision} & 75.000 &  & \multicolumn{1}{c|}{-} & \multicolumn{1}{c|}{full-precision} & 75.000 \\ \cline{1-3} \cline{5-7} 
\end{tabular}
\end{table*}